\documentclass{article}
\usepackage{spconf}
\usepackage{cite}
\usepackage{amsmath,amssymb,amsfonts}
\usepackage{algorithmic}
\usepackage{graphicx}
\usepackage{textcomp}
\usepackage{xcolor}
\usepackage{subfigure}
\usepackage{caption}
\usepackage{multirow, array}

\title{Multitask Learning and Joint Optimization for Transformer-RNN-Transducer Speech Recognition
}

\name{Jae-Jin Jeon \qquad Eesung Kim}

\address{AI R\&D Team, Kakao Enterprise\\ 235, Pangyoyeok-ro, Bundang-gu, Seongnam-si, Gyeonggi-do 13494, Korea \\
\{jeffrey.j, chris.ekim\}@kakaoenterprise.com}
\begin{document}
\maketitle

\begin{abstract}
Recently, several types of end-to-end speech recognition methods named transformer-transducer were introduced. According to those kinds of methods, transcription networks are generally modeled by transformer-based neural networks, while prediction networks could be modeled by either transformers or recurrent neural networks (RNN). This paper explores multitask learning, joint optimization, and joint decoding methods for transformer-RNN-transducer systems. Our proposed methods have the main advantage in that the model can maintain information on the large text corpus. We prove their effectiveness by performing experiments utilizing the well-known ESPNET toolkit for the widely used Librispeech datasets. We also show that the proposed methods can reduce word error rate (WER) by 16.6 \% and 13.3 \% for test-clean and test-other datasets, respectively, without changing the overall model structure nor exploiting an external LM. 

\end{abstract}


\section{Introduction}

In contrast to conventional automatic speech recognition (ASR) systems that comprise acoustic, pronunciation, and language models, end-to-end ASR systems model the whole modules in neural networks and directly output label sequences. Among various end-to-end ASR structures, connectionist temporal classification (CTC),  recurrent neural network transducer (RNN-T), and attention-based encoder-decoder (AED) are widely studied\cite{graves2006}\cite{graves2012}\cite{Chan2015}.

In \cite{graves2012}, RNN-T has been introduced where long short-term memory (LSTM) architectures are adopted for both transcription and prediction networks. The two networks are linked by a joint network to produce output labels, such as a blank symbol. However, various structures such as convolutional neural network (CNN) and transformer have been utilized instead of RNN\cite{han2020contextnet}\cite{Yeh2019}; therefore, we adopted only the transducer terminology omitting RNN in this paper. 

Transformers in \cite{vaswani2017} were introduced to solve various sequence-to-sequence problems and replaced the conventional LSTM-based sequence processing methods. The algorithm effectively attends all the context vectors by adopting multiheaded self-attention. Recently, transformer-transducer ASR systems were proposed based on this structure in \cite{Yeh2019}\cite{Zhang2020}.  As shown in \cite{Yeh2019}, the combination of transformer-based transcription network and LSTM-based prediction network performs best; compared to \cite{Zhang2020}, where only transformers have been adopted for both networks.

Transducer ASR structures have two distinct networks, which are the transcription and prediction networks. The two networks are shallowly linked via the joint network, and the prediction network only inputs the label information. In \cite{rao2017}, they proposed pretraining methods for the transcription and prediction network layers, both based on LSTM. Transcription network layers are pretrained utilizing phoneme, grapheme, and word piece labels with CTC losses, while prediction network layers are pretrained with the text corpus with LM loss. Then, the whole networks are finally fine-tuned using the speech-label datasets. Even though pretraining methods are preferred to random initialization method, there still exist limitations. Pretraining methods cannot fully exploit large scale text corpus information while fine-tuning proceeds, and the CTC and LM classification layers are discarded finally. 

Joint optimization for CTC/AED has successfully introduced and further improved ASR performance via joint decoding in \cite{watanabe2018}. However, it is challenging to integrate LM into the system directly, because the encoder and decoder are deeply entangled, that is, the context vectors from the encoder are transmitted directly to the decoder input layer. 


In this paper, we propose multitask learning and joint optimization for the transformer-RNN-transducer ASR systems to overcome the limitations of conventional methods. Joint optimization with CTC loss on transcription network and LM loss on prediction network impose strong regularization on the networks. Moreover, the LM constructed by the proposed method remains effective regardless of the number of training epochs. Therefore, in addition to the result that the joint optimization procedures themselves contribute to enhancing performance, joint decoding with the LM further reduces WER. The organization of this paper is as follows. Section \ref{sec:proposed} will present the proposed methods, and section \ref{sec:experiment} will explain experiments. Then, we will summarize the result in section \ref{sec:conclusion}.

\section{Proposed system}\label{sec:proposed}
\begin{figure}[t]
\centerline{\includegraphics[width=\linewidth]{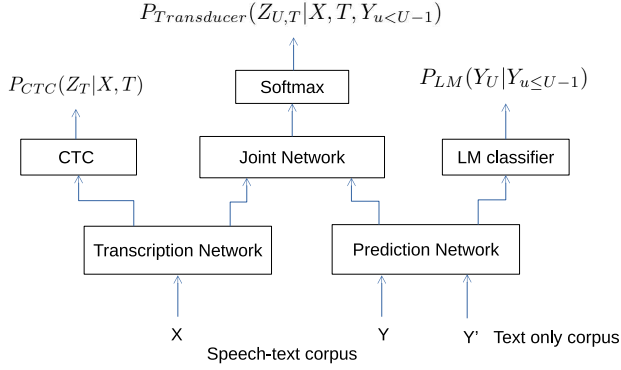}}
\caption{Proposed training structure for joint optimization and multitask learning}
\label{fig:proposed}
\end{figure}

The proposed method is depicted in fig. \ref{fig:proposed}. The difference between the proposed method and the basic transducer-based ASR system is that two classification layers are placed at the end of the transcription and prediction networks, respectively. Last layer of both CTC and LM classifier is a single fully connected (FC) layer followed by softmax function. Assuming that transcription and the prediction vectors denoted as $\mathbf{f}_t \in \mathbb{R}^{D_{TR}}$ and $\mathbf{g}_u \in \mathbb{R}^{D_{PR}}$ are output vectors from the corresponding networks, outputs from the CTC and the LM classifier can be expressed as

\begin{equation*}
    \mathbf{o}_{CTC} = \mathit{Softmax}( \mathbf{f}_t\mathbf{W}_{CTC})
\end{equation*}
and
\begin{equation*}
    \mathbf{o}_{LM} = \mathit{Softmax}( \mathbf{g}_u \mathbf{W}_{LM}),
\end{equation*}
respectively, where  $\mathbf{W}_{CTC} \in \mathbb{R}^{D_{TR}\times D_{CTC}}$ and $\mathbf{W}_{LM} \in \mathbb{R}^{D_{PR}\times D_{LM}}$. Here, dimensions of the LM, the transducer, and the LM outputs are denoted as $\mathbf{o}_{CTC} \in \mathbb{R}^{D_{CTC}}$, $\mathbf{o}_{trans.} \in \mathbb{R}^{D_{trans.}}$, and $\mathbf{o}_{LM} \in \mathbb{R}^{D_{LM}}$.

It should be noted that $D_{CTC}$, $D_{trans.}$, and $D_{LM}$ can or should be different. Assuming that the transducer output labels are trained by sentence piece model (SPM) in \cite{kudo2018sentencepiece}, $D_{LM}$ can be determined by vocabulary size of SPM. For the transducer output, it needs the blank symbol, so that $D_{trans.} = D_{LM} + 1$. Though the CTC's vocabulary size need not be the same as that of the transducer output, in this paper, we choose the same vocabulary size for the CTC and the transducer, that is, $D_{CTC} = D_{trans.}$.


Here, CTC and transducer losses defined respectively in \cite{graves2006} and \cite{graves2012} are denoted as $L_{CTC}$, $L_{trans.}$. The LM loss $L_{LM}$ is negative log probability of the predicted label given the previous sequence.

\subsection{Multitask learning and joint optimization}
The proposed system depends on multitask learning method as it should learn two types of speech recognition models and an additional language model. Moreover, we consider the three losses simultaneously and jointly optimize the total loss function defined as 
\begin{equation}
L_{total} = \alpha_{1} L_{CTC} + \alpha_{2}L_{Trans.} + \alpha_{3} L_{LM}.\label{eq:joint_opt}
\end{equation}


Even though both CTC-based and transducer-based ASR systems adopt a blank symbol, the purpose of using such a symbol is quite different. The blank symbol makes the beam search algorithm finish consuming the current transcription vector and proceed to next frame for transducer-based systems. In other words, the decoder can produce many labels at one frame until it meets a blank symbol. However, in CTC-based systems, each transcription vector is related to only one label: either a blank symbol or a token. Therefore, joint optimization in CTC and transducer losses can induce the decoder to produce one symbol at each frame, which is the same motivation for CTC/AED system in \cite{watanabe2018}.

Joint optimization of transducer and LM requires two different types of data: speech-label corpus and text only corpus. The two types of data cannot be inputted into the system simultaneously. Training procedure utilizing the two corpora will be explained in section \ref{sec:experiment}.

\subsection{Joint decoding}
Under the proposed joint optimization method, the CTC and the LM classifiers remain effective, so that we can utilize them in the decoding process. If we consider all the losses, the best decoding process should find a sequence which maximizes log probability given the input speech, which can be expressed as
\begin{equation}
\begin{aligned}
\hat{Y} =  \operatorname*{argmax}_{Y} [\beta_{1} log P_{CTC}(Y|X) +
\beta_{2} log P_{Trans.}(Y|X)  \\
+ \beta_{3} log P_{LM}(Y)].\label{eq:joint_dec}
\end{aligned}
\end{equation}

 Based on the beam search of the RNN-transducer in \cite{graves2012}, the probability generated from the LM can be easily integrated into the one-pass decoding process. Meanwhile, the decoding process of the CTC is quite different; transducer-based ASR systems generally show better performance than CTC-based systems. Therefore, in this paper, we focus on joint decoding only with the LM, even though it is possible to re-score the sequence probability with the CTC.

\subsection{Transcription network structure}
\begin{figure}[t]
\centerline{\includegraphics[width=0.4\linewidth]{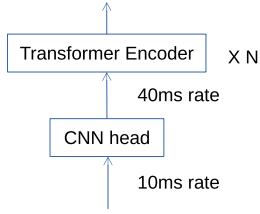}}
\caption{Transcription network structure}
\label{fig:encoder}
\end{figure}

\begin{figure}[t]
\centering
     \begin{subfigure}[]
         \centering
         \includegraphics[width=0.3\linewidth]{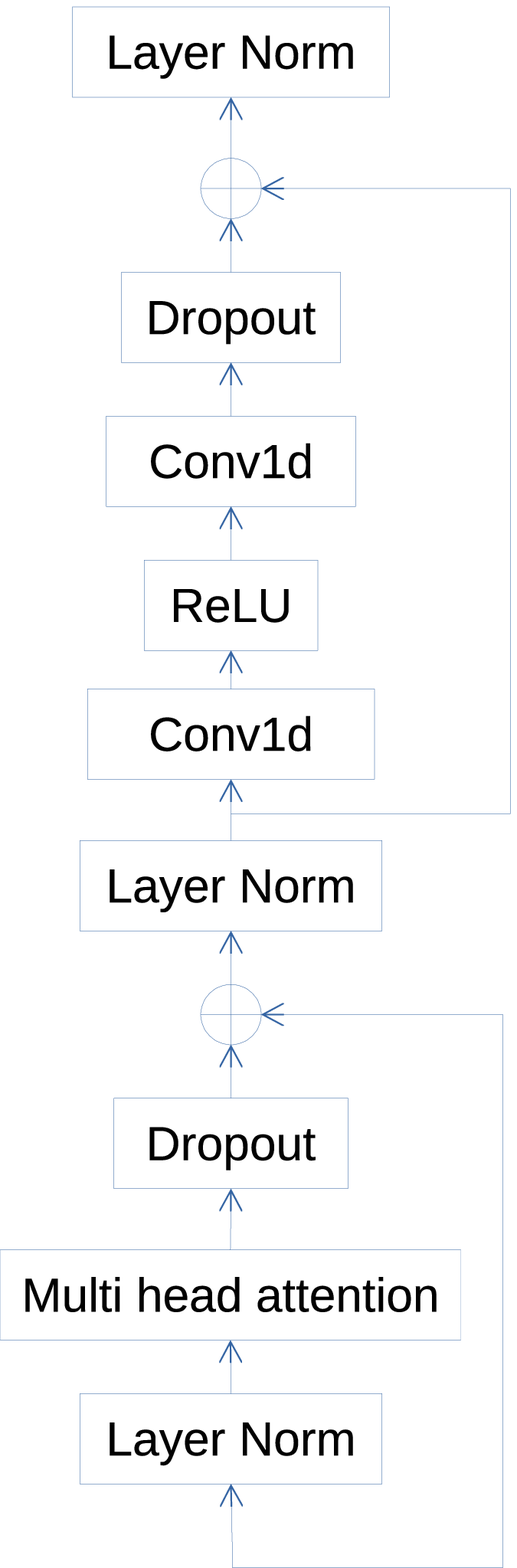}
         \label{fig:transformer}
     \end{subfigure}
     \hspace{0.05\textwidth}
     \begin{subfigure}[]
         \centering
         \includegraphics[width=0.2\linewidth]{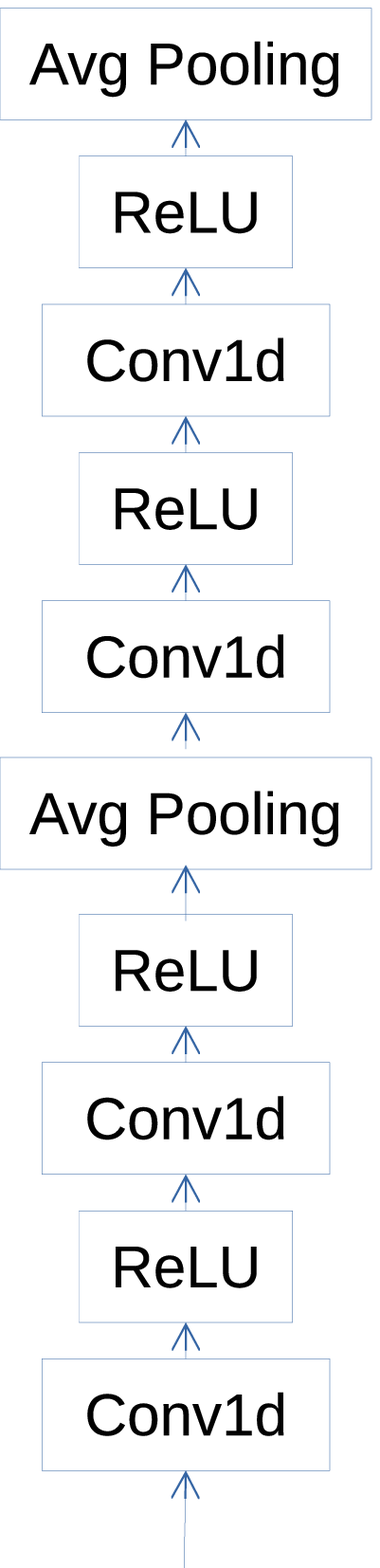}
         \label{fig:cnn_head}
     \end{subfigure}
        \caption{Transformer (a) and CNN head (b) structures}
        \label{fig:encoder_detail}
\end{figure}

The transcription network in this paper comprises one CNN head and multiple layers of transformer encoders, as in fig. \ref{fig:encoder}. The purpose of using CNN head is the same as that of VGGNet in \cite{Yeh2019}, that is, embedding positional information and down-sampling input speech frames. In this paper, we utilize 1D-CNNs instead of 2D-CNNs. Details of the CNN head is depicted in fig. \ref{fig:encoder_detail} (b), where four CNN layers embed positional information into the signal frames, and two average pooling layers with kernel size 2 reduce the frame rate by 4. 

When it comes to additional usage of 1D-CNNs into following transformer layers, there are motivations that CNNs can embed positional information and that the neural networks can focus more on the local features \cite{Mohamed2019}\cite{Yeh2019}\cite{Gulati2020ConformerCT}. Therefore, we adopt the transformer structure of fastspeech in \cite{Ren2019}, where the original feed-forward neural networks are replaced by 1D-CNN layers, as depicted in \ref{fig:encoder_detail} (a).

\subsection{Prediction and joint network}
The prediction network inputs labels encoded by SPM in \cite{kudo2018sentencepiece}. The tokens are embedded into a learnable continuous vector space and the embedded vectors are inputted to LSTM layers.

Ouput vectors for the transducer can be expressed as 
\begin{equation*}
    \mathbf{o}_{trans.} = \mathit{Softmax}( \phi(\mathbf{W}_{TR}\mathbf{f}_t +  \mathbf{W}_{PR}\mathbf{g}_u)\mathbf{W}_{o})
\end{equation*}
where $\mathbf{W}_{TR} \in \mathbb{R}^{D_{TR} \times D_{J}}$, $\mathbf{W}_{PR} \in \mathbb{R}^{D_{PR} \times D_{J}}$, and $\mathbf{W}_o \in \mathbb{R}^{D_J \times D_{trans.}}$. Nonlinear activation $\phi ()$ is hypertangent function.  

\section{Experiments}\label{sec:experiment}
For experiments, we adopted the ESPNET toolkit in \cite{watanabe2018} and modified the training module enabling the proposed multitask learning and joint optimization. Then the publicly available Librispeech dataset in \cite{panayotov2015} is used for training the ASR systems, and WER performances for the test datasets are compared.

Log Mel-filter bank features of dimension 128 are extracted every 10ms and then are normalized with global mean/variance computed from the training dataset. During training, specaugment methods is applied with time warping, frequency mask, and time mask. Time warping parameter $W$, frequency mask parameters $F$:$m_f$, and time mask parameters $T$:$m_t$ in \cite{Park_2019} are set to 5, 32:2, and 40:2, respectively. A sentence piece model with 2048 symbols is trained from transcriptions of the training set and serves as the output symbols. The Adam optimizer in \cite{kingma2017adam} with $\beta_1 = 0.9$, $\beta_2 = 9.98$, $\epsilon = 1e-9$ is utilized for training the models. Warm-up steps are 25k, peak learning rate is $1.0/\sqrt{D_{trans.}}$, and the learning rate diminishes linearly afterwards. During training, we found that the transducer loss function does not correlate with WER performance, so that we trained every model equally for 130 epochs, and model parameters of last 15 epochs are averaged for testing. We utilize a standard beam search with beam size of 20 for decoding.

Parameters for the transcription and the prediction networks are summarized in table \ref{tab:params_trans} and \ref{tab:params_pred}, respectively.

\begin{table}[h]
    \centering
    \begin{tabular}{c|c}
    \hline
     
    Parameter & value\\
    \hline\hline
         CNN head output & 512\\
         CNN head kernel & 3\\
         Average pooling kernel & 2\\
         \hline
         Transformer Layers & 15\\
         Embedding size & 512 \\
         Attention heads & 8 \\
         CNN kernel & 3\\
         First 1D CNN output & 2048\\
         Second 1D CNN output & 512\\
         Dropout rate & 0.1\\
         \hline
         
    \end{tabular}
    \caption{Transcription network parameters}
    \label{tab:params_trans}
\end{table}

\begin{table}[h]
    \centering
    \begin{tabular}{c|c}
    \hline
     
    Parameter & value\\
    \hline\hline
        Vocabulary size of SPM & 2048\\
        SPM embedding dims & 256 \\
         LSTM layers & 2\\
         LSTM cell size & 1024 \\
        Joint network dims & 1024\\
        Dropout rate & 0.1 \\
         \hline
         
    \end{tabular}
    \caption{Prediction and joint network parameters}
    \label{tab:params_pred}
\end{table}

\subsection{Training procedure}
To prove the effectiveness of the proposed methods, three training procedures with \{$\alpha_1, \alpha_2, \alpha_3$\}s in (\ref{eq:joint_opt}) of \{0.5, 0.1, 0.0\}, \{0.0, 1.0, 1.0\}, and \{0.5, 1.0, 1.0\} are conducted. They are joint optimizations based on the combinations of loss functions - joint optimization with CTC, LM, and CTC+LM. Here, it should be noted that two different types of data are used for training the neural networks, that is, speech-label corpus and text-only corpus, which cannot be inputted to the system simultaneously. Moreover, the scales of both the CTC and the transducer losses are similar, whereas that of the LM loss is quite different. Therefore, we follow the next two-step training procedure. 

Firstly, mini-batch data of text corpus is loaded on GPUs, and the model uses the data to calculate gradients on the prediction network and the LM classifier. We do not update the parameters of the model at this step, just applying gradient-clipping to the computed gradients. Next, speech-label data is loaded, and the model calculates the loss function of $\alpha_1 L_{CTC} + \alpha_2 L_{trans.}$. After back-propagation, gradients of the prediction network, CTC, transducer, and the prediction network except the LM classifier are computed at this step. By the two-step training procedure, gradients on the prediction network are calculated two times and accumulated. Finally, after applying gradient clipping again, updating all of the model parameters is conducted. Here, max norm of the gradients is set to 5. 


\subsection{Performance}

The experiment results are presented in table \ref{tab:joint_opt_results}. From now, we use expression of \{test-clean WER, test-other WER\} for the results of the two test datasets. Baseline WERs without any joint optimization are \{4.2, 10.5\}. After utilizing joint optimization only with the CTC loss, the WERs drop to \{3.9, 9.9\}. Considering only the LM loss without the CTC loss, WER scores are \{3.8, 9.9\}. Moreover, joint optimization concerning all of the three losses shows the WERs of \{3.7, 9.5\}. The experiments have shown that joint optimization method can effectively regularize both of transcription and the prediction networks and improve overall system performance without any additional computational burden in the decoding process.

Exploiting an external LM in a one-pass decoding process requires an amount of computational power because it should address all hypotheses of predicted results. However, the proposed method can utilize the internal LM that is consisted of the prediction network and the LM classsifier. If the LM cooperates in the decoding process, the added computational burden is only related to the LM classifier that is a single FC layer. In this experiment, $\{\beta_1, \beta_2, \beta_3\}$ in (\ref{eq:joint_dec}) are set to \{0, 1.0, 0.1\}, where we do not consider joint decoding with the CTC, as mentioned. Joint decoding with the LM also contributes to improving the performance of \{3.5, 9.1\}. As a whole, all the proposed methods can 
reduce the WERs by 16.6\% and 13.3 \% without changing transducer model structures or transformer-RNN-transducer model size.

In \cite{Irie_2019}, LSTM-based LM showed best performance with 4 layers and 2048 cell size. Assuming the proposed method can thus fully exploit the text corpus of Librispeech, there may still exist room for improving performance by increasing prediction network size. Hence, we increased only the number of LSTM layers to 5, and the enlarged prediction network showed WER performance as \{3.4, 8.9\}.

\begin{table}[t]
    \centering
    \begin{tabular}{c|c|c}
    \hline
    \multicolumn{1}{c|}{\multirow{2}{*}{Methods}} &  \multicolumn{2}{c}{WER} \\
    \cline{2-3} 
    & test clean & test other\\
    \hline\hline
    
         Baseline structure &  4.2 & 10.5\\
         \hline
         Joint opt. with CTC & 3.9 & 9.9 \\
         Joint opt. with LM & 3.8 & 9.9\\
         Joint opt. with CTC+LM & 3.7 & 9.5\\
         \hline
         + Joint decoding with LM & 3.5 & 9.1\\
         \hline
         + Large prediction network & 3.4 & 8.9\\
         \hline
         
    \end{tabular}
    \caption{Librispeech dataset results of joint optimization and joint decoding}
    \label{tab:joint_opt_results}
\end{table}

\section{Conclusion}\label{sec:conclusion}
In this paper, we adopted multitask learning, joint optimization, and joint decoding for the transformer-RNN-transducer ASR system. The joint optimization of the three loss functions enhanced system performance without increasing inference computational complexity. Also, joint decoding reduces recognition error with an increase of relatively small computational burden for the LM classifier. The proposed method can be applied to any transcription and prediction network that constitutes transducer-loss-based ASR systems.


\bibliographystyle{./bibliography/IEEEtran}
\bibliography{my_bib}

\begin{thebibliography}{10}
\providecommand{\url}[1]{#1}
\csname url@samestyle\endcsname
\providecommand{\newblock}{\relax}
\providecommand{\bibinfo}[2]{#2}
\providecommand{\BIBentrySTDinterwordspacing}{\spaceskip=0pt\relax}
\providecommand{\BIBentryALTinterwordstretchfactor}{4}
\providecommand{\BIBentryALTinterwordspacing}{\spaceskip=\fontdimen2\font plus
\BIBentryALTinterwordstretchfactor\fontdimen3\font minus
  \fontdimen4\font\relax}
\providecommand{\BIBforeignlanguage}[2]{{%
\expandafter\ifx\csname l@#1\endcsname\relax
\typeout{** WARNING: IEEEtran.bst: No hyphenation pattern has been}%
\typeout{** loaded for the language `#1'. Using the pattern for}%
\typeout{** the default language instead.}%
\else
\language=\csname l@#1\endcsname
\fi
#2}}
\providecommand{\BIBdecl}{\relax}
\BIBdecl

\bibitem{graves2006}
A.~Graves, S.~Fern{\`a}ndez, F.~Gomez, and J.~Schmidhuber, ``Connectionist
  temporal classification: Labelling unsegmented sequence data with recurrent
  neural networks,'' in \emph{Proceedings of the International Conference on
  Machine Learning, ICML 2006}, 2006.

\bibitem{graves2012}
A.~Graves, ``Sequence transduction with recurrent neural networks,''
  \emph{arXiv preprint arXiv:1211.3711}, 2012.

\bibitem{Chan2015}
W.~Chan, N.~Jaitly, Q.~V. Le, and O.~Vinyals, ``Listen, attend and spell,''
  \emph{arXiv preprint arXiv:1508.01211}, 2015.

\bibitem{han2020contextnet}
W.~Han, Z.~Zhang, Y.~Zhang, J.~Yu, C.-C. Chiu, J.~Qin, A.~Gulati, R.~Pang, and
  Y.~Wu, ``Contextnet: Improving convolutional neural networks for automatic
  speech recognition with global context,'' \emph{arXiv preprint
  arXiv:2005.03191}, 2020.

\bibitem{Yeh2019}
C.-F. Yeh, J.~Mahadeokar, K.~Kalgaonkar, Y.~Wang, D.~Le, M.~Jain, K.~Schubert,
  C.~Fuegen, and M.~L. Seltzer, ``Transformer-transducer: End-to-end speech
  recognition with self-attention,'' \emph{arXiv preprint arXiv:1910.12977},
  2019.

\bibitem{vaswani2017}
A.~Vaswani, N.~Shazeer, N.~Parmar, J.~Uszkoreit, L.~Jones, A.~N. Gomez,
  L.~Kaiser, and I.~Polosukhin, ``Attention is all you need,'' \emph{arXiv
  preprint arXiv:1706.03762}, 2017.

\bibitem{Zhang2020}
Q.~Zhang, H.~Lu, H.~Sak, A.~Tripathi, E.~McDermott, S.~Koo, and S.~Kumar,
  ``Transformer transducer: A streamable speech recognition model with
  transformer encoders and rnn-t loss,'' \emph{arXiv preprint
  arXiv:2002.02562}, 2020.

\bibitem{rao2017}
K.~{Rao}, H.~{Sak}, and R.~{Prabhavalkar}, ``Exploring architectures, data and
  units for streaming end-to-end speech recognition with rnn-transducer,'' in
  \emph{2017 IEEE Automatic Speech Recognition and Understanding Workshop
  (ASRU)}, 2017, pp. 193--199.

\bibitem{watanabe2018}
S.~Watanabe, T.~Hori, S.~Karita, T.~Hayashi, J.~Nishitoba, Y.~Unno, N.~{Enrique
  Yalta Soplin}, J.~Heymann, M.~Wiesner, N.~Chen, A.~Renduchintala, and
  T.~Ochiai, ``Espnet: End-to-end speech processing toolkit,'' in
  \emph{Interspeech}, 2018, pp. 2207--2211.

\bibitem{kudo2018sentencepiece}
T.~Kudo and J.~Richardson, ``Sentencepiece: A simple and language independent
  subword tokenizer and detokenizer for neural text processing,'' \emph{arXiv
  preprint arXiv:1808.06226}, 2018.

\bibitem{Mohamed2019}
A.~Mohamed, D.~Okhonko, and L.~Zettlemoyer, ``Transformers with convolutional
  context for {ASR},'' \emph{arXiv preprint arXiv:1904.11660}, 2019.

\bibitem{Gulati2020ConformerCT}
A.~Gulati, J.~Qin, C.-C. Chiu, N.~Parmar, Y.~Zhang, J.~Yu, W.~Han, S.~Wang,
  Z.~Zhang, Y.~Wu, and R.~Pang, ``Conformer: Convolution-augmented transformer
  for speech recognition,'' \emph{arXiv preprint arXiv:2005.08100}, 2020.

\bibitem{Ren2019}
Y.~Ren, Y.~Ruan, X.~Tan, T.~Qin, S.~Zhao, Z.~Zhao, and T.-Y. Liu, ``Fastspeech:
  Fast, robust and controllable text to speech,'' in \emph{NeurIPS}, 2019.

\bibitem{panayotov2015}
V.~{Panayotov}, G.~{Chen}, D.~{Povey}, and S.~{Khudanpur}, ``Librispeech: An
  asr corpus based on public domain audio books,'' in \emph{2015 IEEE
  International Conference on Acoustics, Speech and Signal Processing
  (ICASSP)}, 2015, pp. 5206--5210.

\bibitem{Park_2019}
D.~S. Park, W.~Chan, Y.~Zhang, C.-C. Chiu, B.~Zoph, E.~D. Cubuk, and Q.~V. Le,
  ``Specaugment: A simple data augmentation method for automatic speech
  recognition,'' \emph{Interspeech 2019}, Sep 2019.

\bibitem{kingma2017adam}
D.~P. Kingma and J.~Ba, ``Adam: A method for stochastic optimization,''
  \emph{arXiv preprint arXiv:1412.6980}, 2014.

\bibitem{Irie_2019}
K.~Irie, A.~Zeyer, R.~Schlüter, and H.~Ney, ``Language modeling with deep
  transformers,'' \emph{Interspeech 2019}, Sep 2019.

\end{thebibliography}
\end{document}